\documentclass{article}
\usepackage{microtype}
\usepackage{graphicx}
\usepackage{subcaption}
\usepackage{booktabs}
\usepackage{hyperref}

\usepackage[accepted]{icml2026}
\usepackage{amsmath}
\usepackage{amssymb}
\usepackage{mathtools}
\usepackage{amsthm}
\usepackage[capitalize,noabbrev]{cleveref}

\icmltitlerunning{Memory-Efficient Partitioned DNN Inference on Resource-Constrained Android Crowds}

\begin{document}
\twocolumn[
  \icmltitle{Memory-Efficient Partitioned DNN Inference\\
             on Resource-Constrained Android Crowds}

  \icmlsetsymbol{equal}{*}

  \begin{icmlauthorlist}
    \icmlauthor{Lakshani Manamperi}{equal,uom}
    \icmlauthor{Thiwanka Pathirana}{equal,uom}
    \icmlauthor{Disumi Pathirana}{equal,uom}
    \icmlauthor{Nipun Premarathna}{equal,uom}
    \icmlauthor{Kutila Gunasekera}{uom}
  \end{icmlauthorlist}

  \icmlaffiliation{uom}{Department of Computer Science and Engineering,
    University of Moratuwa, Moratuwa, Sri Lanka}

  \icmlcorrespondingauthor{Lakshani Manamperi}{lakshani.21@cse.mrt.ac.lk}

  \icmlkeywords{memory-efficient inference, model partitioning, edge ML,
    ONNX, mobile crowd computing, deployment-aware scheduling,
    compressed tensor transport, JIT loading, Android, TinyML}

  \vskip 0.3in
]
\printAffiliationsAndNotice{\icmlEqualContribution}

\begin{abstract}
Deploying large deep neural networks on memory-constrained mobile devices
is a central challenge in edge ML.
While compression, pruning, and quantization reduce per-parameter cost,
transformer-based models remain too large for the 3.3--7.4\,GB RAM envelope
of commodity Android handsets.
We present the \emph{DNN pipeline scheduling subsystem} of CROWDio, which
achieves practical ONNX inference across resource-constrained Android workers
\emph{without} model modification, by distributing memory pressure across
devices via five mechanisms: JIT deferred partition loading, a
single-partition-resident constraint, a 4-tier affinity scheduler, a
zlib-compressed tensor transport, and a streaming 1:1 dependency model.
Evaluated on DistilBERT~\citep{sanh2019distilbert} (${\approx}$67\,M
parameters, SST-2) across five Android handsets over ten runs, our system
holds peak per-device RSS to 43$\pm$2\,MB and limits battery draw to
50$\pm$3\,mAh per run, while streaming concurrency cuts batch latency
34\% below barrier synchronisation.
\end{abstract}

\section{Introduction}
\label{sec:intro}

Resource-efficient ML uses compression, pruning, quantization, and
distillation to reduce inference footprints.
A complementary, under-explored axis is \emph{deployment-aware
partitioning}: rather than shrinking a model, distribute its layers across
multiple memory-constrained devices so the aggregate fleet memory
accommodates the full model while no single device exceeds its RAM
budget~\citep{li2020pipeswitch,huang2019gpipe}.

This is especially relevant in \emph{Mobile Crowd Computing}
(MCdC)~\citep{pramanik2024}, where volunteer Android handsets contribute
idle compute.
Commodity handsets carry 3.3--7.4\,GB RAM; a mid-sized transformer such
as DistilBERT requires 2--4\,GB for a single ONNX session, leaving
negligible headroom for the OS and background processes.
Existing solutions---pruning~\citep{han2016deep}, post-training
quantization~\citep{jacob2018quantization}, distillation---still assume a
single-device host and require model modification.
Split computing systems (SPINN~\citep{laskaridis2020spinn},
Neurosurgeon~\citep{kang2017neurosurgeon}) target a single, fixed partition
point to a reliable cloud backend; neither addresses the stochastic
availability and heterogeneous RAM of a volunteer crowd, nor the need to
schedule dynamically across more than two pipeline stages.

We present the \textbf{DNN pipeline scheduling subsystem} of
CROWDio~\citep{crowdio_anon}, contributing:
\begin{itemize}\setlength\itemsep{0pt}
  \item \textbf{JIT deferred loading} (\cref{sec:pipeline_model}): shards
        loaded only when tasks are pending, bounding peak RAM.
  \item \textbf{Single-partition residency} (\cref{sec:pipeline_model}):
        one ONNX session per worker, hard-capping RSS.
  \item \textbf{4-tier affinity scheduler} (\cref{sec:scheduling}):
        prioritises resident $\to$ cached $\to$ idle $\to$ evict, cutting
        cold-load latency on the critical path.
  \item \textbf{Compressed tensor transport} (\cref{sec:transport}):
        zlib-encoded self-describing payloads with filesystem fallback.
  \item \textbf{Streaming 1:1 dependency model} (\cref{sec:streaming}):
        per-input task independence replaces batch barriers.
\end{itemize}

\section{Background and Related Work}
\label{sec:background}

\textbf{On-device inference.}
Quantization~\citep{jacob2018quantization}, pruning~\citep{han2016deep},
and distillation~\citep{hinton2015distilling} are orthogonal to our
approach: CROWDio loads compressed ONNX shards unchanged.

\textbf{Model partitioning.}
GPipe~\citep{huang2019gpipe} and PipeDream~\citep{narayanan2019pipedream}
partition layers across GPU workers.
Melon~\citep{chen2022melon}---the closest prior art---offloads layers
within a single device; our JIT strategy (\cref{sec:pipeline_model})
extends this to a heterogeneous fleet.
EdgePipe~\citep{zhao2022edgepipe} shows affinity-aware placement cuts
inter-stage transfer by 43\%, motivating \cref{sec:scheduling}.
Shah et~al.~\citep{shah2023edgebased} establish codec-compressed activation
transport for split inference, informing \cref{sec:transport}.

\textbf{MCdC infrastructure.}
Hyrax~\citep{marinelli2009hyrax} and Misco~\citep{dou2010misco} established
MapReduce on smartphone clusters.
Nagesh et~al.~\citep{nagesh2025} contribute dependency-aware DAG
work-stealing that informs our dependency manager
(\cref{sec:streaming,sec:flow}).

\section{Memory-Aware Partition Scheduling}
\label{sec:sysarch}

\subsection{System Overview}

\begin{figure}[ht]
  \vskip 0.1in
  \begin{center}
    \centerline{\includegraphics[width=\columnwidth]{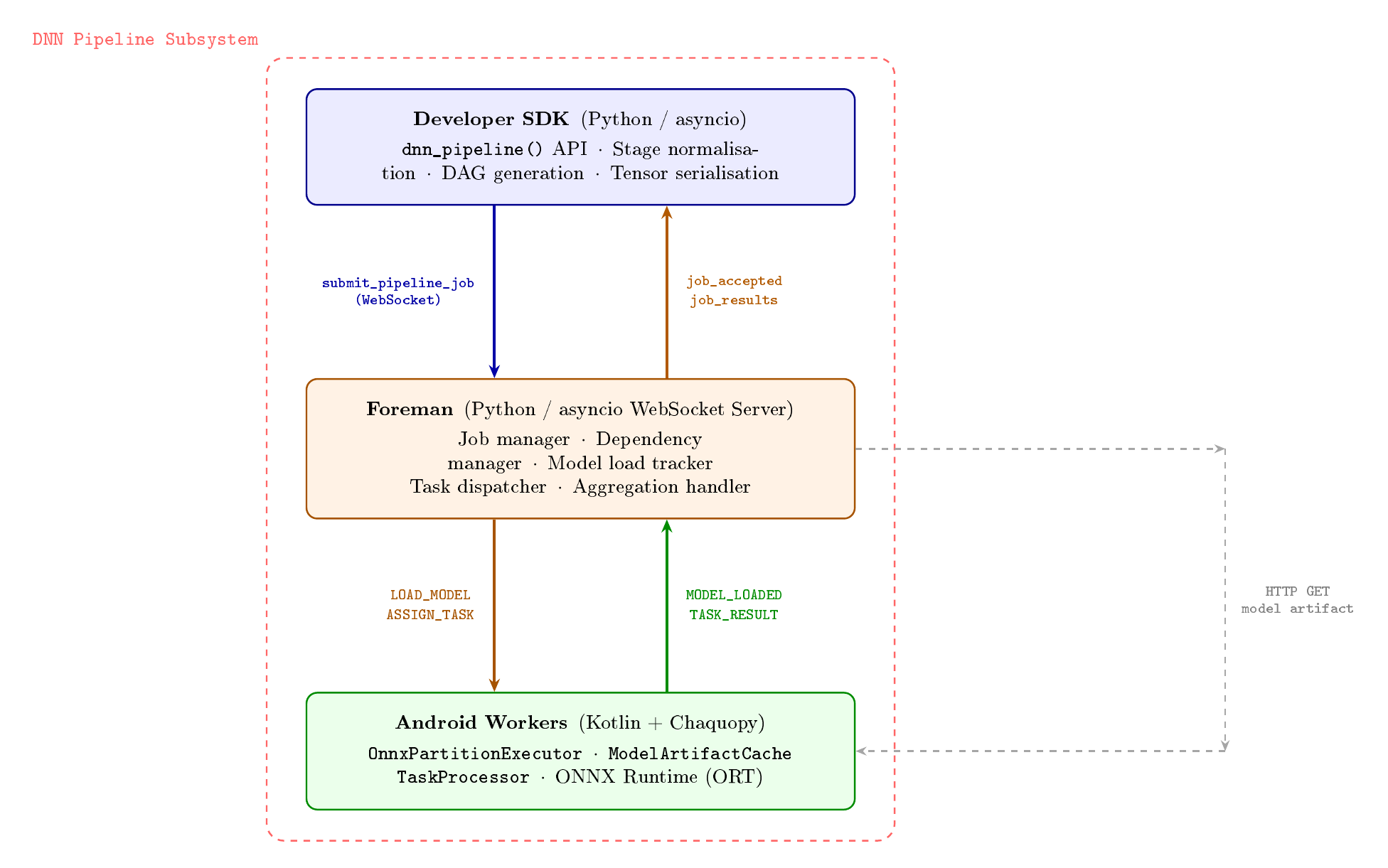}}
    \caption{CROWDio three-layer architecture. The SDK submits jobs; the
             Foreman orchestrates scheduling and failure recovery; Workers
             execute inference under the single-residency constraint.}
    \label{fig:sysarch}
  \end{center}
  \vskip -0.2in
\end{figure}

CROWDio has three layers over persistent WebSocket
connections~\citep{fette2011websocket} (\cref{fig:sysarch}).
The \textbf{Developer SDK} generates a linear DAG from ordered ONNX
artefacts and submits SHA-256-checksummed model blobs (see \cref{sec:flow}).
The \textbf{Foreman} handles scheduling (\cref{sec:scheduling}) and
failure recovery.
\textbf{Android Workers} execute ONNX inference under the single-residency
constraint (\cref{sec:pipeline_model}) and report telemetry every 30\,s.

\subsection{Pipeline Model and Memory Budget}
\label{sec:pipeline_model}

A DNN is split into $S$ ordered stages, each an independent ONNX artefact
assigned to a distinct worker.
Our reference workload is DistilBERT~\citep{sanh2019distilbert} for SST-2
sentiment analysis (${\approx}$67\,M parameters, 6 encoder layers) split
at layer~3 into three cells:

\noindent\emph{Stage~0 (\texttt{cell\_a}, Embedding + Layers~0--2).}
Input: $[B,S]$ integer tensor; output: $[B,S,H]$ hidden states.
Eagerly broadcast to all workers as the smallest shard, ensuring every
device is immediately ready to accept Stage~0 tasks on job arrival.

\noindent\emph{Stage~1 (\texttt{cell\_b}, Layers~3--5).}
JIT-loaded only after the first Stage~0 output arrives, ensuring the
largest shard is never held idle in RAM.

\noindent\emph{Stage~2 (\texttt{cell\_c}, Pre-classifier + Classifier).}
Likewise JIT-loaded on first Stage~1 completion.
Output: $[B,2]$ logits averaged across inputs for the majority prediction.

The \textbf{single-partition-resident} constraint limits each worker to one
active ONNX session at a time.
Combined with JIT loading, peak RSS is bounded to one shard's footprint
regardless of pipeline depth.
A monolithic DistilBERT ONNX session requires 2--4\,GB~\citep{sanh2019distilbert};
our scheme holds per-device RSS to 43$\pm$2\,MB (\cref{sec:eval}), as no
worker ever holds more than one partition simultaneously.
Partition correctness is verified end-to-end: chained ONNX Runtime output
vs.\ full PyTorch output yields max-abs error $\approx 0$.

\subsection{Streaming Dependency Model}
\label{sec:streaming}

\begin{figure}[ht]
  \vskip 0.1in
  \begin{center}
    \centerline{\includegraphics[width=\columnwidth]{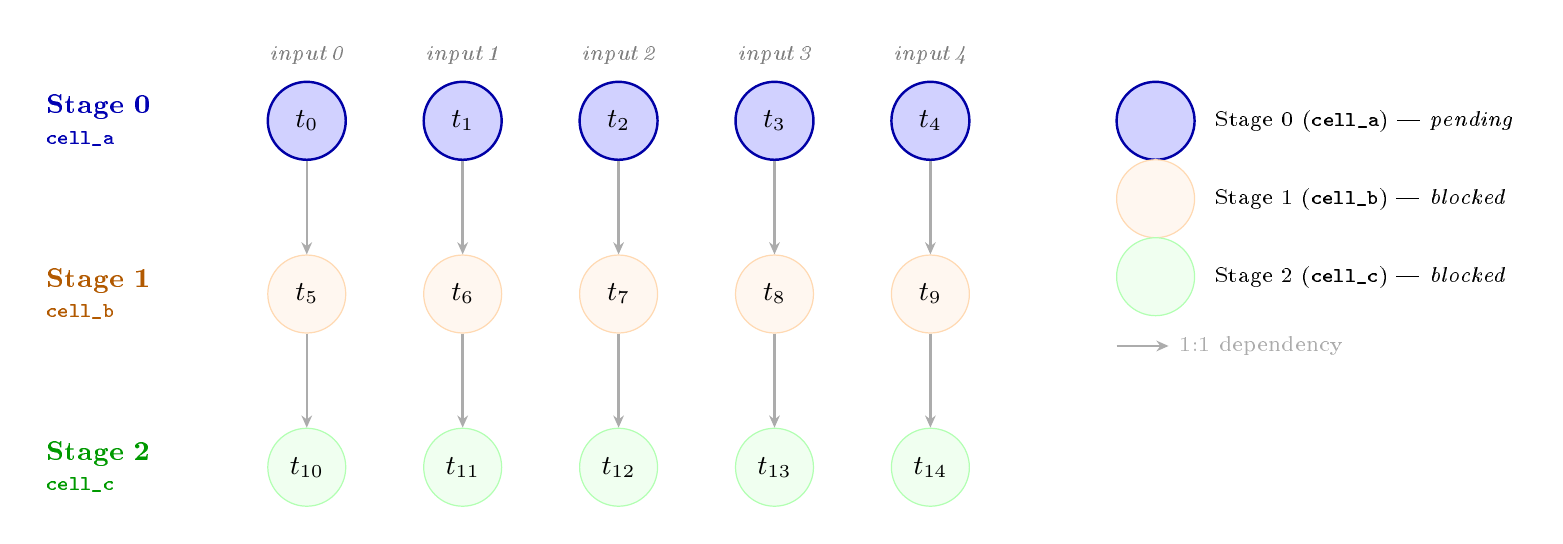}}
    \caption{Task graph for the 3-stage pipeline (streaming, $N{=}5$).
             Stage~0 tasks $t_0$--$t_4$ are initially pending; each
             Stage~1/2 task carries a per-input 1:1 dependency, yielding
             maximal concurrency.}
    \label{fig:pipeline}
  \end{center}
  \vskip -0.2in
\end{figure}

For $N$ inputs and $S$ stages the Foreman materialises $N{\times}S$ task
records.
\textbf{Streaming mode} assigns each Stage-$k$ task a 1:1 dependency on
its Stage-$(k{-}1)$ predecessor for the same input $i$, decoupling
per-input progress and enabling maximal concurrency (\cref{fig:pipeline};
\cref{tab:taskgraph}).
\textbf{Barrier mode} holds all Stage-$k$ tasks until every
Stage-$(k{-}1)$ task finishes (1:$N$ dependency), amplifying straggler
effects on heterogeneous hardware~\citep{wei2013scheduling}; the latency
penalty is quantified in \cref{sec:eval}.
For the reference workload ($N{=}5$, $S{=}3$), streaming yields 15
independent tasks versus a synchronisation barrier at every stage
transition---mirroring PipeDream's micro-batching
rationale~\citep{narayanan2019pipedream} adapted to heterogeneous mobile
workers.

\begin{table}[t]
  \caption{Task graph: 3 stages $\times$ 5 inputs (streaming mode).}
  \label{tab:taskgraph}
  \vskip -0.05in
  \begin{center}
    \begin{small}
      \begin{sc}
        \begin{tabular}{@{}lccc@{}}
          \toprule
          Task IDs & Stage & State & Depends On \\
          \midrule
          $t_0$--$t_4$       & 0 & pending & --- \\
          $t_5$--$t_9$       & 1 & blocked & $t_{0..4}$ (1:1) \\
          $t_{10}$--$t_{14}$ & 2 & blocked & $t_{5..9}$ (1:1) \\
          \bottomrule
        \end{tabular}
      \end{sc}
    \end{small}
  \end{center}
  \vskip -0.15in
\end{table}

\subsection{Compressed Tensor Transport}
\label{sec:transport}

Intermediate activations are encoded as self-describing JSON payloads
(\texttt{dtype}, \texttt{shape}, \texttt{compression}, base64-zlib byte
string) embedded over the existing WebSocket channel, with symmetric
serialisation in Python (SDK/Foreman) and Kotlin (Android worker).
This achieves 62$\pm$4\% compression on $[1,768]$ FP32 tensors
(3\,072 raw bytes), reducing inter-stage transfer to ${\approx}$1\,168
bytes per tensor on average.
The protocol is codec-agnostic: LZ4 or Snappy may be substituted for
lower-latency deployments without changes to the framing schema.

When compressed payload size exceeds threshold $\tau_{\mathrm{ws}}$
(default: 1\,MB), the payload is written to a shared filesystem store and
only a reference key is forwarded to the Foreman, avoiding double-transfer
through the broker (\cref{tab:payloadstore}).
This two-path design ensures the WebSocket channel is never saturated by
large intermediate tensors, which is critical under Wi-Fi contention in
a multi-device crowd.

\begin{table}[t]
  \caption{Activation payload routing.}
  \label{tab:payloadstore}
  \vskip -0.05in
  \centering
  \begin{small}
    \begin{sc}
      \begin{tabular}{@{}p{1.4cm}p{2.7cm}p{2.3cm}@{}}
        \toprule
        Condition & Worker output & Worker input \\
        \midrule
        ${\leq}\tau_{\mathrm{ws}}$ &
          Inline in message &
          Decoded from field \\[2pt]
        ${>}\tau_{\mathrm{ws}}$ &
          Write to fs-store; send ref key &
          Read from fs-store \\
        \bottomrule
      \end{tabular}
    \end{sc}
  \end{small}
  \vskip -0.1in
\end{table}

\subsection{4-Tier Model Affinity Scheduler}
\label{sec:scheduling}

The affinity scheduler wraps any base ranking algorithm (FIFO,
EDAS~\citep{edas2017}, ARAS~\citep{aras2010}, MABAC~\citep{mabac2015})
with two pre-dispatch gates.

\textbf{Model gating} restricts dispatch to workers whose required
partition is already in session memory, preventing the race condition where
inference arrives before a JIT-loaded partition is ready
(\cref{sec:pipeline_model}).

\textbf{Affinity promotion} ranks eligible workers by residency tier
(\cref{tab:worker_select}).
Crucially, Tier~4 is not an unconstrained fallback: it triggers an explicit
\texttt{UNLOAD\_MODEL} of the resident shard followed by a fresh
\texttt{LOAD\_MODEL}, incurring the highest latency while still preserving
the single-residency invariant.
EdgePipe~\citep{zhao2022edgepipe} reports 43\% inter-stage transfer
reduction from equivalent affinity strategies, validating this design.

Within a tier, workers are ranked by live heartbeat telemetry (CPU, RAM,
battery, RTT, temperature) weighted via Shannon
entropy~\citep{shannon1948,pramanik2021mcdm}: criteria that are uniform
across workers receive near-zero weight; high-variance criteria dominate.
A two-phase batch assignment resolves contention: worker-driven claiming
(rarest-coverage-first among Tier-1 workers) precedes task-driven MCDM
fill for remaining assignments, ensuring that workers already hosting a
shard are fully utilised before eviction is considered.

\begin{table}[t]
  \caption{Worker selection tiers for deferred partition load.}
  \label{tab:worker_select}
  \vskip -0.05in
  \centering
  \begin{small}
    \begin{sc}
      \begin{tabular}{@{}clp{3.0cm}@{}}
        \toprule
        Tier & Strategy & Criterion \\
        \midrule
        1 & Resident & Partition in session memory \\
        2 & Cached   & Partition file on-disk \\
        3 & Idle     & No resident partition \\
        4 & Evict    & Unload current, then load \\
        \bottomrule
      \end{tabular}
    \end{sc}
  \end{small}
  \vskip -0.1in
\end{table}

\section{End-to-End Execution}
\label{sec:flow}

\textbf{SDK submission.}
The developer specifies ordered ONNX stage artefacts and execution mode.
The SDK performs stage normalisation, auto-topology generation (linear DAG
with source/intermediate/sink classification), acyclicity validation, and
submits a \texttt{SUBMIT\_PIPELINE\_JOB} message with base64-encoded,
SHA-256-checksummed model blobs.

\begin{figure}[ht]
  \vskip 0.1in
  \begin{center}
    \centerline{\includegraphics[width=\columnwidth]{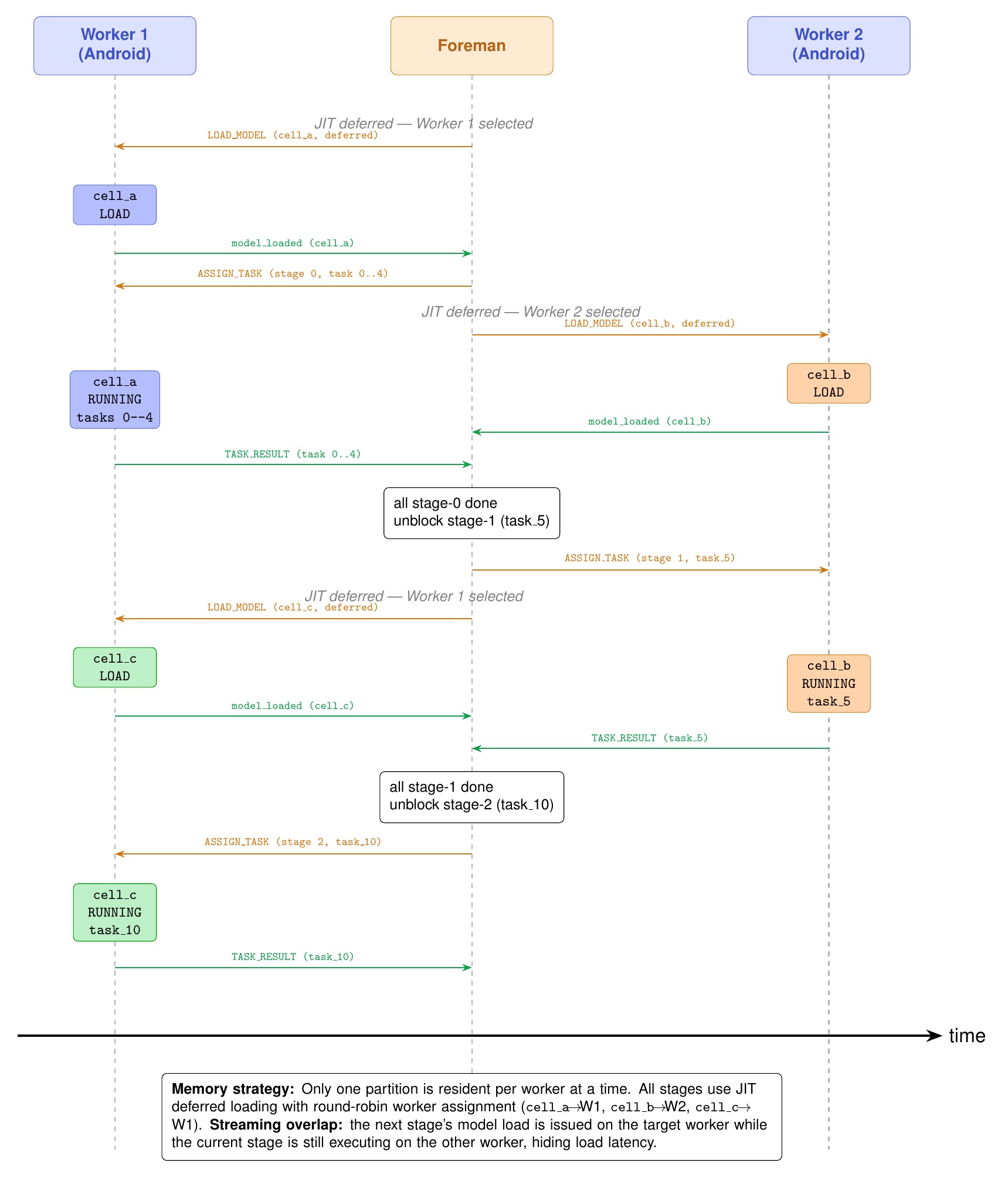}}
    \caption{Partition loading sequence. \texttt{cell\_a} is eagerly
             broadcast; \texttt{cell\_b}/\texttt{cell\_c} are JIT-loaded
             only after the first upstream task completes.}
    \label{fig:modelload}
  \end{center}
  \vskip -0.2in
\end{figure}

\textbf{Foreman job creation.}
The Foreman materialises $N{\times}S$ task records and registers the DAG
per \cref{sec:streaming}.
\texttt{cell\_a} is eagerly broadcast to all workers; JIT load instructions
for \texttt{cell\_b}/\texttt{cell\_c} are stored and dispatched only on
the first upstream task completion, enforcing the memory budget invariant
(\cref{fig:modelload}).

\textbf{Worker inference.}
The worker decodes the input activation (inline or from the filesystem
store per \cref{sec:transport}), runs inference via a session-cached
\texttt{OrtSession}, and re-encodes the output as a compressed payload.
Sessions are cached per model path and released on \texttt{UNLOAD\_MODEL}.

\textbf{Pipeline progression.}
The dependency manager injects the upstream payload into the downstream
task, decrements its dependency counter to zero, and enqueues it for the
affinity selector (\cref{sec:scheduling}).
On all-task completion, logit scores across inputs are averaged and the
majority-class prediction returned to the SDK.

\textbf{Failure recovery.}
On worker disconnect, the dynamic topology planner scores surviving workers
by success rate, RAM, battery, GPU availability, and residency tier, and
re-dispatches affected tasks without interrupting in-flight inferences.

\section{Evaluation}
\label{sec:eval}

\textbf{Setup.}
Five heterogeneous Android handsets (3.3--6\,GB RAM, Android~13--14,
ONNX Runtime~1.17) over local Wi-Fi run the DistilBERT SST-2 pipeline
($N{=}5$, $S{=}3$); all figures are per-device mean\,$\pm$\,std over ten
independent runs.
We compare \textbf{CROWDio Streaming} (\cref{sec:streaming}) against
\textbf{CROWDio Barrier} (\cref{tab:eval}).

\textbf{Peak RAM.}
The single-partition-resident constraint (\cref{sec:pipeline_model}) holds
per-device RSS to 43$\pm$2\,MB across all handsets, including the two
lowest-RAM devices (3.3\,GB).
A monolithic DistilBERT ONNX session requires 2--4\,GB~\citep{sanh2019distilbert},
making single-device deployment infeasible on resource-constrained handsets
without partitioning.

\textbf{Pipeline latency.}
Streaming mode completes the $N{=}5$ batch in 18.4$\pm$1.1\,s---a
\textbf{34\% reduction} over barrier mode (27.9$\pm$2.3\,s), as 1:1
per-input dependencies eliminate cross-input straggler blocking.

\textbf{Load time and battery.}
Cold-start load completes in 48$\pm$4\,s; Tier-2 cache hits
(\cref{sec:scheduling}) reduce this to 6$\pm$1\,s on warm runs.
Battery draw is 50$\pm$3\,mAh per device per run, with session
initialisation dominating over inference.

\textbf{Compression.}
The zlib transport achieves 62$\pm$4\% compression on DistilBERT
hidden-state activations ($[1,768]$ FP32 tensors), reducing inter-stage
transfer from 3\,072 to ${\approx}$1\,168 bytes per tensor.

\begin{table}[t]
  \caption{Per-device results: 5-device Android crowd, DistilBERT SST-2,
           $N{=}5$, mean\,$\pm$\,std over 10 runs.}
  \label{tab:eval}
  \vskip -0.05in
  \centering
  \setlength{\tabcolsep}{5pt}
  \begin{small}
    \begin{sc}
      \begin{tabular}{@{}lrr@{}}
        \toprule
        Metric & Streaming & Barrier \\
        \midrule
        Peak RAM (RSS)   & \textbf{43$\pm$2\,MB}    & 43$\pm$2\,MB    \\
        Pipeline latency & \textbf{18.4$\pm$1.1\,s} & 27.9$\pm$2.3\,s \\
        Cold-start load  & 48$\pm$4\,s              & 48$\pm$4\,s     \\
        Warm load        & \textbf{6$\pm$1\,s}      & 6$\pm$1\,s      \\
        Battery delta    & \textbf{50$\pm$3\,mAh}   & 53$\pm$4\,mAh   \\
        Compression      & \multicolumn{2}{c}{62$\pm$4\%}              \\
        \bottomrule
      \end{tabular}
    \end{sc}
  \end{small}
  \vskip -0.1in
\end{table}

\section{Discussion and Conclusion}
\label{sec:conclusion}

We presented the DNN pipeline scheduling subsystem of CROWDio, enabling
multi-stage ONNX inference of DistilBERT across resource-constrained
Android workers without pruning, quantization, or cloud back-ends.
JIT loading and single-residency (\cref{sec:pipeline_model}) bound peak RAM
to one shard's footprint, making deployment feasible on handsets where a
monolithic session (2--4\,GB) would starve the OS.
Streaming concurrency (\cref{sec:streaming}) eliminates cross-input
straggler blocking, cutting batch latency 34\% below barrier mode
(\cref{tab:eval}).
The core insight is that \emph{deployment-aware partitioning} and
\emph{per-shard compression} are complementary: partitioning distributes
memory pressure across the fleet while compression reduces each shard's
transfer and storage footprint.

Several trade-offs merit attention.
The single-residency constraint is conservative on higher-RAM ($>$6\,GB)
devices where holding two shards simultaneously would still leave OS
headroom; memory-adaptive multi-partition residency is a natural extension.
The 48\,s cold-start cost---covering partition download and ONNX session
initialisation---is significant for latency-sensitive applications, but
collapses to 6\,s on Tier-2 warm runs (\cref{sec:scheduling}); aggressive
pre-fetching informed by job arrival patterns is a natural mitigation.
zlib compression adds CPU-bound serialisation on the critical path; LZ4
or Snappy may be preferable for lower-latency deployments.
Auto-topology generation is currently restricted to linear chains; branch
or merge topologies require developer-supplied explicit graphs~\citep{moritz2018ray}.

Future work will combine memory-adaptive residency, non-linear topology
support, and energy-aware scheduling~\citep{pramanik2024energy} with
quantization-aware shard training to push transformer-class inference to
sub-2\,GB RAM devices.

\section*{Impact Statement}
This work lowers the barrier for transformer inference in low-resource
settings by enabling commodity Android handsets to participate in inference
without model modification.
Volunteer-crowd deployments introduce battery and user-consent
considerations that practitioners should address via transparent opt-in
mechanisms.

\bibliography{crowdio_dnn}
\bibliographystyle{icml2026}

\end{document}